\pdfoutput=1

\documentclass[11pt]{article}

\usepackage[preprint]{acl}

\usepackage{times}
\usepackage{latexsym}
\usepackage{graphicx}

\usepackage[T1]{fontenc}

\usepackage[utf8]{inputenc}

\usepackage{microtype}

\usepackage{inconsolata}
\usepackage{amsmath}
\usepackage{amsfonts}
\usepackage[framemethod=TikZ]{mdframed}
\usepackage{tikz}
\usepackage{color,colortbl}
\usepackage{xspace}
\usepackage{algorithm}
\usepackage{multirow}
\usepackage{enumerate}
\usepackage{makecell}
\usepackage{arydshln}
\usepackage{pifont}
\usepackage{algpseudocode}
\usepackage{booktabs}
\usepackage{tablefootnote}
\usepackage{bbding}
\usepackage{ulem}
\usepackage{cite}
\usepackage{dsfont}
\usepackage{listings}

\usepackage{subcaption}

\usepackage{adjustbox}
\usepackage{xcolor} 
\usepackage[most, breakable]{tcolorbox}
\definecolor{lightgray}{RGB}{249, 249, 249}
\newtcolorbox{mybox}{
  breakable,
  arc=3mm,
  colback=lightgray,
  colframe=black,
  boxrule=0.3mm,
}

\newtcolorbox{mybox2}{
  breakable,
  arc=3mm,
  colback=white,
  colframe=black,
  boxrule=0.3mm,
}

\definecolor{blueish}{RGB}{250, 250, 255}
\definecolor{greenish}{RGB}{200, 255, 200}
\definecolor{ForestGreen}{RGB}{64, 136, 39}
\definecolor{redish}{RGB}{212, 57, 57}
\definecolor{highlight}{RGB}{175, 255, 100}
\definecolor{darkred}{RGB}{139, 0, 0}
\definecolor{gray95}{gray}{0.05}
\definecolor{rowgray}{RGB}{224, 224, 224}

\newmdenv[
    tikzsetting= {fill=blueish},
    skipabove=0.33em,
    skipbelow=0.33em,
    linewidth=1pt,
    innerleftmargin=4pt,
    innerrightmargin=4pt,
    innertopmargin=2pt,
    innerbottommargin=2pt,
    linecolor=gray95,
    roundcorner=2pt, 
    shadowsize=4pt,
    shadowcolor=gray95
]{answerbox}

\newenvironment{result}
{\begin{answerbox}}
{\end{answerbox}}

\title{Rethinking Repetition Problems of LLMs in Code Generation}

\author{\textbf{Yihong Dong},
    \textbf{Yuchen Liu},
    \textbf{Xue Jiang},
  \textbf{Zhi Jin}, 
   and \textbf{Ge Li}\footnotemark[1]\\
   Key Laboratory of High Confidence Software Technologies (Peking University), \\ Ministry of Education; School of Computer Science, Peking University, Beijing, China \\
    \texttt{\{dongyh, liuyuchen1, jiangxue\}@stu.pku.edu.cn}, 
    \texttt{\{zhijin, lige\}@pku.edu.cn} \\ 
}

\begin{document}

\maketitle

\begin{abstract}
With the advent of neural language models, the performance of code generation has been significantly boosted. However, the problem of repetitions during the generation process continues to linger. Previous work has primarily focused on content repetition, which is merely a fraction of the broader repetition problem in code generation. A more prevalent and challenging problem is structural repetition. In structural repetition, the repeated code appears in various patterns but possesses a fixed structure, which can be inherently reflected in grammar. In this paper, we formally define structural repetition and propose an efficient decoding approach called RPG, which stands for Repetition Penalization based on Grammar, to alleviate the repetition problems in code generation for LLMs. Specifically, RPG first leverages grammar rules to identify repetition problems during code generation, and then strategically decays the likelihood of critical tokens that contribute to repetitions, thereby mitigating them in code generation. To facilitate this study, we construct a new dataset CodeRepetEval to comprehensively evaluate approaches for mitigating the repetition problems in code generation. Extensive experimental results demonstrate that RPG substantially outperforms the best-performing baselines on CodeRepetEval dataset as well as HumanEval and MBPP benchmarks, effectively reducing repetitions and enhancing the quality of generated code.
\footnote{Our code and dataset are 
 avaliable at \url{https://github.com/LYC127/RPG}}
\end{abstract}

\section{Introduction}
Code generation seeks to automatically produce code that aligns with user intents, which is a research hotspot in the fields of artificial intelligence, natural language processing, and software engineering \citep{CodeRL, codet, model_previous_3}. In recent years, the emergence of neural language models has shown remarkable advancements in code generation \citep{codex, GPT-4}. However, even well-trained large language models (LLMs) may suffer from repetition problems, which hurts the code generation quality of LLMs substantially \citep{Hallucination24}.

\begin{figure*}[h!]
\centering
\includegraphics[width=1.05\textwidth]{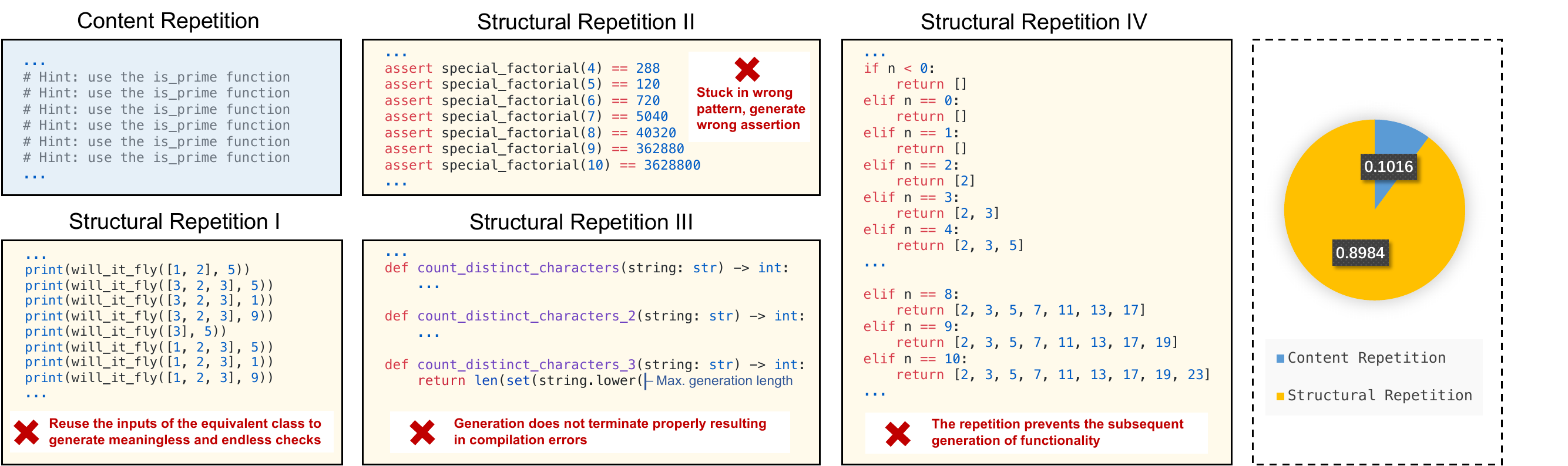}
\caption{Examples of repetition problems in code generation, collected from the well-trained LLMs, e.g., CodeLlama \citep{codellama} and ChatGPT \citep{ChatGPT} (\textbf{Left}). The statistical percentage of two repetition forms occurs in the generated code of LLMs (\textbf{Right}).}
\label{example}
\end{figure*}

Recent studies about repetition problems of LLMs are primarily focused on content repetition \citep{Rep22, Rep23}, which refers to the results of the generation system always containing duplicate fragments \citep{Rep21}. 
However, our preliminary investigation of LLM's repetition problem in code generation reveals that content repetition constitutes only a minor portion of them, as shown in Figure \ref{example}. In contrast, a predominant form of repetition in the generated results involves the repeated occurrence of similar codes with fixed structural patterns, which we term `structural repetition'\footnote{Generally, in code generation, content repetition can be regarded as a special type of structural repetition. However, in this paper, structural repetition is defined to exclude content repetition for clarity and distinction.}. Distinct from content repetitions, the pattern of different structural repetitions varies markedly (e.g., structural repetitions I-IV), making structural repetitions hard to detect and handle. Given the diversity and complexity of structural repetitions, previous approaches tailored for content repetitions are insufficient to address them effectively. 
Therefore, it is necessary and significant to explore the structural repetitions in code generation. 

In this paper, we propose an effective decoding approach RPG: \textbf{R}epetition \textbf{P}enalization based on \textbf{G}rammar, to alleviate repetition problems of LLMs in code generation. Considering different code fragments with the same structural patterns can be represented by identical grammar rules, RPG employs the pushdown automaton built on grammar rules to detect repetition problems during the generation process, and then strategically decreases the likelihood of key tokens that contribute to repetitions. 
RPG offers two main benefits: 1) it curtails the endless generation of meaningless, repetitive code, thereby saving tokens and time-consuming; 2) it realigns the LLMs' generation back to the correct generation path, enhancing the quality of code generation.
Moreover, we construct a new dataset, named CodeRepetEval, for evaluating approaches to mitigate the repetition problems in code generation. 
Extensive experimental results and analyses verify the effectiveness and generality of RPG.

Our main contribution can be summarized as fourfold. 1) We first formally define structural repetitions, which are more prevalent than content repetitions in code generation. 2) We present RPG, a novel decoding approach that leverages pushdown automaton to identify and mitigate repetitive problems in code generation from grammar perspective. 3) We construct CodeRepetEval dataset covering three scenarios, with data derived from artificial synthesis, code generation benchmarks, and real-world repositories, to facilitate subsequent research for repetitive problems in code generation. 4) RPG substantially outperforms the best-performing baselines in various scenarios, which alleviates both structural and content repetitions, and achieves better code generation quality. 

\section{Motivation Example}
\begin{figure*}[h!]
\centering
\includegraphics[width=0.88\textwidth]{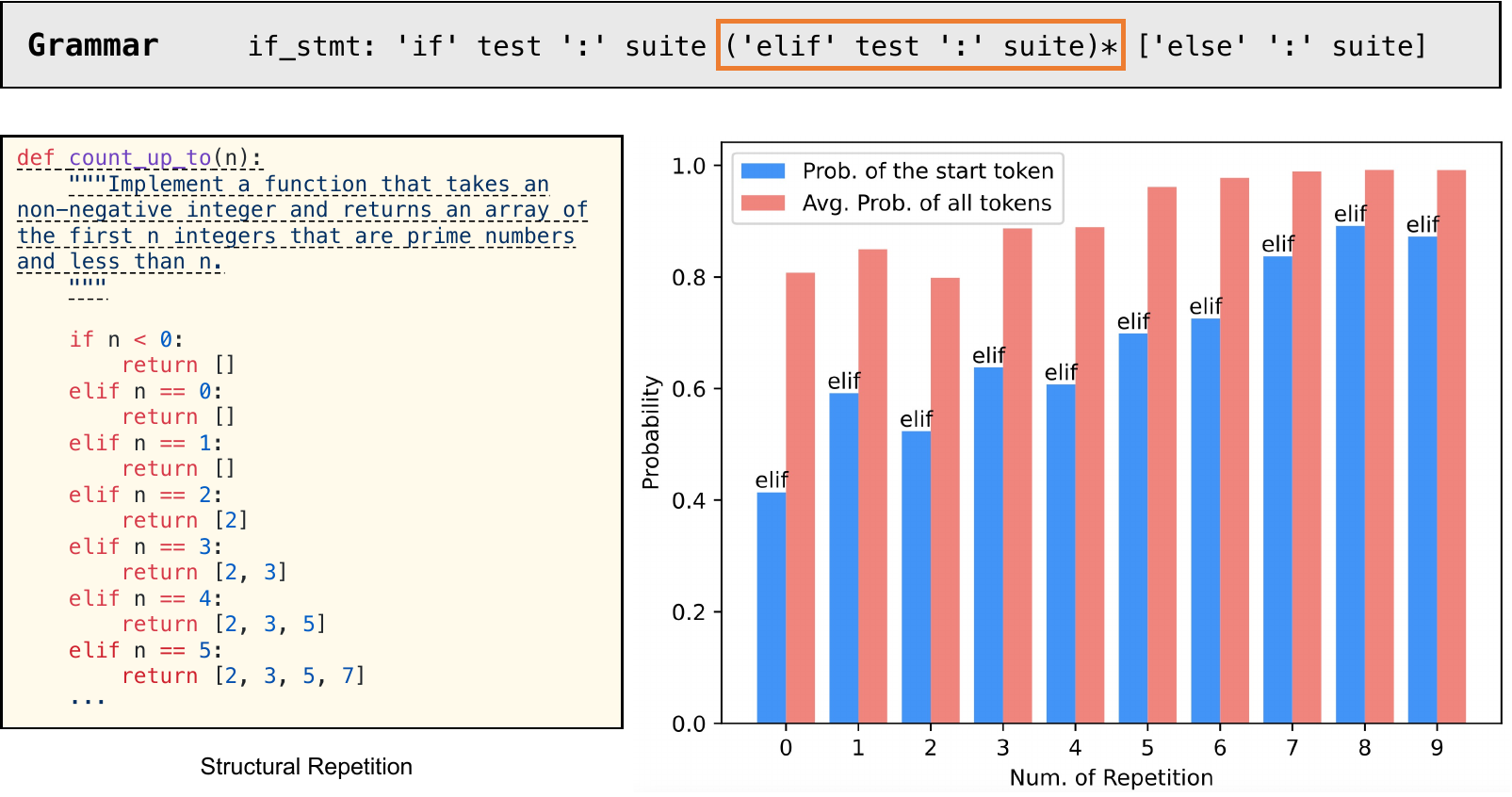}
\caption{A case of structural repetition generated by CodeLlama with temperature = 0, where the
\dashuline{dashed-underline text} is the prompt (\textbf{Left}). The corresponding grammar rules of structural patterns (\textbf{Top}). LLM's probabilities of generated tokens in each repetition (\textbf{Right}).}
\label{motivation}
\end{figure*}

The repetition problem in code generation remains an underexplored challenge, usually resulting in redundancy and errors. Figure \ref{motivation} showcases an example where LLMs generate the code containing structural repetitions. This generated code is plagued by repetitions with the fixed structural pattern starting with `elif'. In each repetition, LLMs generate different conditions following the start token `elif' and varying statements under these conditions. Despite the content of each repetition differing, both the probability of the start token and the average probabilities of all tokens in each repetition exhibit an upward trend as the number of repetitions increases, showing a self-reinforcement effect as the right side of Figure \ref{motivation}.
This phenomenon means that the start token in each repetition will serve as an anchor point. As the model continues to generate code, it relies on this anchor, reinforcing its choice and making it increasingly difficult to diverge from the structural repetition.
This ultimately leads to the subsequent generation getting stuck in endless and meaningless (or even erroneous) repetitions. 

According to principles of compilation \citep{alfred2007compilers}, we discover that massive patterns of structural repetition are inherently reflected in grammar, i.e., the potential positions where code can be repeated are determined by explicit grammar rules.
For example, the structural repetitions of the code in Figure \ref{motivation} adhere to \textit{(`elif' test `:' suite)*} within \textit{if\_stmt} of grammar, where $*$ denotes that its preceding expression can be repeated zero or more times. 
Although the grammar rules impose no limit on the number of repetitions, human-written code does not repeat endlessly, with the higher the number of repetitions, the lower the likelihood of their occurrence.
Therefore, the prediction confidence of further repetition ought to decrease with a growing number of repetitions, rather than exhibiting the self-reinforcement usually observed in LLMs.

In this paper, we first formally define structural repetitions, and propose RPG to effectively detect and alleviate them in code generation, thereby enhancing the quality of generated code for LLMs.

\section{Definition of Structural Repetition}
Given a token sequence \( X = [x_1, x_2, \cdots, x_{|X|}] \), where \( x_i \) denotes the \( i \)-th token and $|X|$ represents the length of $X$. We denote \( X_{p:q} = [x_p, x_{p+1}, \cdots, x_q] \), where \( (1 \leq p < q \leq |X|) \), as a continuous subsequence of \( X \). Given a mapping function $G$ to represent the underlying structure of $X$, the mapped sequence $\hat{R} = G(X)$ is obtained by applying $G$ to $X$, where $\hat{R} = [\hat{r}_1, \hat{r}_2, \cdots, \hat{r}_{|\hat{R}|}]$. In this paper, \( G \) indicates the context-free grammar\footnote{Programming languages (such as Python, Java, C++, etc.) belong to context-free languages, which means that they are governed by context-free grammars.}, which can be defined as a quad tuple \((N, \Sigma, P, S) \), where \( N \) is a set of non-terminal symbols, \( \Sigma \) is a set of terminal symbols, representing the basic symbols of the language, \( P \) is a set of production rules, with each rule in the form \( A \rightarrow \beta \), where \( A \in N \) and \( \beta \) is a sequence of elements from \( N \cup \Sigma \), and \( S \in N \) is the start symbol, used to begin derivations of strings.

Specifically, the generated sequence $X$ will be reduced in accordance with $G$, ensuring that the reduction of $X$ does not exceed the statement levels of grammar rules, which includes twelve simple statements and nine compound statements\footnote{For instance, simple statements contain `expr\_stmt', `return\_stmt', `raise\_stmt', `import\_stmt', `assert\_stmt', etc. Compound statements contain `if\_stmt', `while\_stmt', `for\_stmt', `try\_stmt', `funcdef', `classdef', etc. Detailed descriptions can be found in Appendix \ref{Full Grammar specification}.}. Thus, the patterns of structural repetitions within $X$ can be defined as:
\begin{align}
    \operatorname{SR}(X) = \{\hat{R}_{p:q} |& \exists 1 \leq p < q \leq |\hat{R}| - q + p,\notag \\ & \forall i \in [p, q], \hat{r}_{i+q-p} = \hat{r}_i\},
    \label{SR}
\end{align}
where structural repetitions exist in $X$, if $\operatorname{SR}(X) \neq \emptyset$, and the elements in $\operatorname{SR}(X)$ represent the patterns of structural repetitions. For example, if $X$ denotes the generated code in Figure \ref{motivation}, then $\hat{R}$ would be ``\textit{$\cdots$ `elif' test `:' suite `elif' test `:' suite $\cdots$}''. Thus, $\hat{R}_{p:q}$ can be the subsequence of $\hat{R}$, i.e., ``\textit{`elif' test `:' suite}'', where $\forall i \in [p, q], \hat{r}_{i+q-p} = \hat{r}_i$. Note that for the same input, \( G \) has a unique output, i.e., if content repetitions exist in \( X \), the repetitions will be preserved in \( G(X) \) as well, which implies content repetitions can also be detected by Eq. \eqref{SR}.

Structural repetition negatively impacts code generation in the following two ways: 1) it fails to terminate properly, rendering the code uncompilable. As repetition persists, the generation of LLMs becomes meaningless or incorrect gradually; 2) it disrupts the generation of code, leading to the absence of a portion of functionality, thereby severely hurting the quality of generated code.

\section{Methodology}
In this section, we will introduce RPG in detail, including three parts, i.e., Reduction to Grammar Rules (\S~\ref{Reduction to Grammar Rules}), Detection of Repetition (\S~\ref{Detection of Repetition}), and Penalization of Repetition (\S~\ref{Penalization of Repetition}).

\subsection{Reduction to Grammar Rules}
\label{Reduction to Grammar Rules}
In problems of structural repetition, while the forms of repeated statements vary, their underlying grammar rules demonstrate similarities.
Following the previous work \citep{CODEP}, we employ the pushdown automaton (PDA) for the reduction of generated codes into their underlying grammar rules during code generation\footnote{We adapt PDA to accommodate the Byte-Pair Encoding (BPE) \citep{BPE} tokenization used by LLMs.  Detailed descriptions can be found in Appendix \ref{PDA4LLM}.}. A PDA can be defined as a seven tuple \( (Q, \Sigma, \Gamma, \delta, q_0, Z, F) \), where \( Q, \Sigma, \Gamma \) are finite sets representing the states, input symbols, and stack symbols, respectively. \( q_0 \) is the initial state, \( z_0 \) is the initial stack symbol, and \( A \) is the set of accepting states, with \( q_0 \in Q \), \( z_0 \in \Gamma \), and \( A \subseteq Q \). \( \delta: Q \times (\Sigma \cup \{\epsilon\}) \times \Gamma \rightarrow \mathcal{P}(Q \times \Gamma^*) \) is the transition function, where \(\epsilon\) denotes the empty string, and \( \Gamma^* \) represents all sequences of stack symbols.

Based on $\delta$ of PDA, we have
\begin{equation}
    q_{t}, z_{t} = \delta(q_{t-1}, x_{t}, z_{t-1}),
\end{equation}
where $q_{t}$ is the state of $t$-th time step, $z_{t}$ is the stack symbol of $t$-th time step, and $x_{t}$ is the generated token of $t$-th time step. According to $q_{t}$ and $z_{t}$, $x_{t}$ can be reduced to its corresponding grammar rule uniquely, which is expressed as: 
\begin{equation}
    \hat{x}_{t} = g(x_t) = [q_{t}, z_{t}], 
\end{equation}
We merge the same adjacent parts in $\hat{X}_{1:t} = [\hat{x}_1, \hat{x}_2, \cdots, \hat{x}_{t}]$ to get the final reduction sequence of grammar rules for $x_{t}$. 
\begin{equation}
    \hat{R}_{1:t} = \operatorname{merge}(\hat{X}_{1:t}), 
\end{equation}
For example, multiple adjacent generated tokens in Figure \ref{motivation} belong to the same terms such as `test' and `suite', the adjacent ones will be merged into a single entity for each term.

\subsection{Detection of Repetition}
\label{Detection of Repetition}

Considering that once repetition problems arise in code generation, they tend to persist until the end of the generation process, we need to detect these repetition problems as they emerge during code generation. 
To detect the repetitions in $\hat{R}_{1:t}$, we employ suffix arrays and longest common prefix (LCP) arrays, which are efficient with the time complexity of $O(n\log n)$ and the space complexity of $O(n)$. The pseudo-code for suffix arrays and the LCP array is provided in Appendix \ref{pseudocode}.

\paragraph{Suffix Array:} The suffix array is an array of integers representing the starting indices of the suffixes of \( \hat{R}_{1:t} \), sorted in lexicographical order. Thus, \( \text{Suf}[i] \) points to the starting index of the \( i \)-th smallest suffix in \( \hat{R}_{1:t} \).

\paragraph{Longest Common Prefix Array:} The LCP array is defined such that \( \text{LCP}[i] \) is the length of the longest common prefix between suffixes starting at \( \text{Suf}[i-1] \) and \( \text{Suf}[i] \) for all \( 1 \leq i < n \) (with \( \text{LCP}[0] \) typically set to 0 for convenience).

Using the LCP array, we identify all positions of $\hat{R}_{1:t}$ where \( \text{LCP}[i] > 0 \). These positions indicate the presence of repetitions of length at least \( \text{LCP}[i] \). Therefore, the repetition patterns within  $X_{1:t}$ can be expressed as:
\begin{align}
    &\text{Rep}(X_{1:t}) = \{\hat{R}_{\text{Suf}[i]: \text{Suf}[i] + \text{LCP}[i]} \ | \ \forall i \in [1, t-1], \notag \\ &\text{LCP}[i] > 0 \ \land \ \text{Suf}[i-1] = \text{Suf}[i] + \text{LCP}[i] \},
\end{align}

\subsection{Penalization of Repetition}
\label{Penalization of Repetition}
Given the identified repetitions, RPG applies a penalization mechanism to discourage the model from generating them in future outputs. The penalization mechanism integrates into the code generation model's scoring function, modifying how tokens are weighted during the generation process.

\paragraph{Dynamic Weight Adjustment:} We define a dynamic weight function, \( \text{Pn}(\cdot) \), which applies a decreasing factor to the score of a token based on the frequency and recency of its associated grammar rule in the sequence \( \hat{R}_{1:t} \). The weight for each token is adjusted as follows:
\begin{equation}
    \text{Pn}(x_t | x_{<t}) = \lambda^{\text{Count}(\text{Rep}(X_{1:t}))},
\end{equation}
where $x_{<t}$ represents $X_{1:t-1}$, \( \lambda \) is a decay factor between 0 and 1, and \( \text{Count}(\text{Rep}(X_{1:t})) \) is the count of times the repetition patterns in \( \text{Rep}(X_{1:t}) \) has appeared. This exponential decay effectively reduces the likelihood of selecting tokens associated with repetitive grammar rules.

\paragraph{Token Scoring Adjustment:} During the code generation, each token's score is recalculated by incorporating the dynamic weight:
\begin{equation}
    s'(x_t| x_{<t}) = s(x_t| x_{<t}) \cdot \text{Pn}(x_t | x_{<t}),
\end{equation}
where \( s(x_t) \) is the original score of the token \( x_t \) provided by the model, and \( g(x_t) \) is the grammar rule associated with \( x_t \). The adjusted score \( s'(x_t) \) influences the token selection process, guiding the model toward less repetitive and more diverse code generation. 

Finally, our RPG approach can be defined as:
\begin{equation}
    \text{RPG}(x_t| x_{<t}) = \arg\max_{x_t} s'(x_t| x_{<t})
\end{equation}

\section{Experiment Setup}
In this section, we will provide the setups of our experiments below. The detailed description of experiment setups can be found in Appendix \ref{app_setup}.

\subsection{Datasets} 
\label{Datasets}
Considering the absence of datasets for repetition problems in code generation, we dedicate more than 400 hours to constructing and examining \textbf{CodeRepetEval} dataset. We simulate repetition problems in code generation covering three scenarios, including artificial synthesis, code generation benchmarks, and real-world repositories. Specifically, \textbf{Artificial Synthesis} scenario involves 512 test samples. Each sample consists of a correct code concatenated with its last repetition patterns 5 to 10 times. \textbf{Code Generation Benchmarks} scenario comprising 512 test samples, which are selected from the generated repetitive codes of three LLMs (i.e., CodeLlama \citep{codellama}, DeepSeek Coder \citep{DeepSeek_Coder}, CodeGen \citep{codegen}, and ChatGPT \citep{ChatGPT}) on HumanEval and MBPP benchmarks. \textbf{Real-world Repositories} scenario includes 1024 test samples, picked from the partial code in real-world repositories, which is identified to induce repetition problems in the generated outputs of the aforementioned LLMs. We employ CodeRepetEval to assess the effectiveness of RPG for addressing repetition problems in code generation scenarios.

Moreover, We also involve four public benchmarks to evaluate the RPG's performance in code generation, including \textbf{HumanEval} \citep{codex}, \textbf{MBPP} \citep{mbpp}, as well as their extended version \textbf{HumanEval-ET} and \textbf{MBPP-ET} \citep{CodeScore}.

\subsection{Baselines}

As our approach is based on decoding that does not require modification and training of the model, we compare it to the four most commonly used decoding approaches, including \textbf{Greedy Sampling}, \textbf{Topk Sampling} \citep{topk}, \textbf{Temperature Sampling} \citep{caccia2019language}, \textbf{Topp Sampling} \citep{Holtzman}. Furthermore, we also compare two representative baselines for addressing content repetition in text generation, including \textbf{Repetition Penalty} \citep{CTRL} for the decoding phase and \textbf{Repetition Dropout} \citep{Rep23} applied on the training phase.
These baselines follow the settings in their original paper.

\subsection{Metrics}
\label{Metrics}
We mainly use six metrics to evaluate approaches for addressing the repetition problems in code generation. \textbf{End-of-sentence Generation Percentage (EGP)} quantifies the frequency with which a model successfully interrupts repetitive sequences to conclude generation, which is determined by calculating the proportion of end-of-sentence (EOS) tokens across all samples generated by the model. \textbf{TR-N} is calculated to measure structural repetitions within a generated sequence at phrase-level, which is defined as $1.0-\frac{|\{G(x)'|\exists p \in [1, |G(x)| - n+1], G(x)'=G(x)_{p:p+n-1}\}|}{|G(x)|-n+1}$, where $|G(x)|$ means the number of elements in $G(x)$. It effectively quantifies the proportion of duplicate n-grams present in its underlying grammar rules, and $n = 4$ in this paper. In contrast, \textbf{TR-S} measures structural repetitions within a generated sequence at statement-level, which is defined as $1.0 - \frac{| \text{unique statements in } G(x) |}{| \text{statements in } G(x) |}$. \textbf{Compiler Correctness Percentage (CCP)}  evaluates whether the generated code is compilable, which is measured by the proportion of code samples that successfully compile. We use \textbf{Time} and \textbf{GenLen} to verify the approaches' efficiency, which means the average time of model generation and the average length of model-generated outputs, respectively.

For HumanEval(-ET) and MBPP(-ET) benchmarks which contain the test cases, we employ \textbf{Pass@k} \citep{alphacode} metric to measure the functional correctness of the generated code by executing test cases.

\subsection{Implementation Details} 
\label{Implementation Details}
In this paper, all experiments are conducted on an A6000 GPU (48GB). We employ CodeLlama-7B as our base model. The decay factor $\lambda$ for the penalization of repetition in RPG is set at 0.9 by default. 
The maximum token length of each approach is set to 1024 in all datasets and scenarios, except for CodeRepetEval (real-world repository) setting to 4096. 
Following the previous work \citep{codex, codellama}, the default temperature for the baselines is set at 0.8. To mitigate the instability of the model sampling, we report the average results of five trials in the experiments.

\begin{table}[th!]
\centering
\caption{Comparison of RPG with commonly used decoding approaches and representative content repetition baselines on CodeRepetEval dataset in three scenarios. 
}
\label{table1}
\resizebox{0.5\textwidth}{!}{
\begin{tabular}{@{}lcccccccc@{}}
\toprule
\multirow{2}{*}{Approach} & \multicolumn{6}{c}{CodeRepetEval}                                                     \\ \cmidrule(l){2-7} 
    & EGP  $\uparrow$ & TR-N $\downarrow$ & TR-S $\downarrow$ & CCP $\uparrow$  & Time $\downarrow$ & \multicolumn{1}{c}{GenLen} \\ \midrule
\multicolumn{8}{c}{\cellcolor{gray!15}\textbf{Code Generation Benchmarks}} \\
Rep\_Penalty       & 0.721          & 0.374          & 0.425          & 0.413          & 16.88          & 689          \\
Rep\_Dropout & 0 & 0.569 & 0.536 & 0.218 & 35.65 & 1024 \\
\hdashline

Greedy                    & 0              & 0.598          & 0.637          & 0.455          & 33.87          & 1024         \\
Temp (t=0.1)              & 0.007          & 0.599          & 0.622          & 0.39           & 37.72          & 1019         \\
Temp (t=0.2)              & 0.03           & 0.603          & 0.63           & 0.413          & 36.13          & 950          \\
Temp (t=0.8)              & 0.578          & 0.423          & 0.441          & 0.433          & 27.13          & 755          \\
Topk (k=5)                & 0.536          & 0.465          & 0.484          & 0.394          & 20.61          & 763          \\
Topk (k=10)               & 0.547          & 0.441          & 0.464          & 0.421          & 29.11          & 752          \\
Topk (k=30)               & 0.628          & 0.415          & 0.443          & 0.442          & 33.99          & 737          \\
Topp (p=0.8)              & 0.046          & 0.559          & 0.549          & 0.379          & 35.17          & 995          \\
Topp (p=0.9)              & 0.102          & 0.53           & 0.512          & 0.391          & 42.03          & 966          \\
Topp  (p=0.95)            & 0.114          & 0.508          & 0.518          & 0.399          & 30.99          & 959          \\ \hdashline
RPG (Ours)                & \textbf{0.912} & \textbf{0.352} & \textbf{0.391} & \textbf{0.805} & \textbf{13.68} & \textbf{565} \\ 
\bottomrule
\multicolumn{8}{c}{\cellcolor{gray!15}\textbf{Artificial Synthesis}} \\
Rep\_Penalty       & 0.679          & 0.624 & 0.521 & 0.467          & 20.91          & 615          \\
Rep\_Dropout & 0 & 0.713 & 0.597 & 0.188 & 32.89 & 1024 \\
\hdashline

Greedy                    & 0              & 0.807          & 0.789          & 0.459          & 40.45          & 1024         \\
Temp (t=0.1)              & 0.016          & 0.798          & 0.785          & 0.452          & 40.16          & 1010         \\
Temp (t=0.2)              & 0.018          & 0.798          & 0.768          & 0.438          & 39.95          & 982          \\
Temp (t=0.8)              & 0.522          & \textbf{0.613}          & \textbf{0.519}          & 0.433          & 23.30          & 675          \\
Topk (k=5)                & 0.503          & 0.659          & 0.558          & 0.45           & 24.35          & 723          \\
Topk (k=10)               & 0.552          & 0.636          & 0.524          & 0.48           & 22.42          & 661          \\
Topk (k=30)               & 0.561          & 0.628          & 0.521          & 0.475          & 22.00          & 653          \\
Topp (p=0.8)              & 0.097          & 0.752          & 0.691          & 0.431          & 37.27          & 946          \\
Topp (p=0.9)              & 0.146          & 0.724          & 0.658          & 0.501          & 35.82          & 915          \\
Topp  (p=0.95)            & 0.201          & 0.681          & 0.637          & 0.454          & 33.19          & 868          \\ \hdashline
RPG (Ours)                & \textbf{0.871} & 0.618 & 0.556 & \textbf{0.731} & \textbf{17.37} & \textbf{489} \\ \bottomrule
\multicolumn{8}{c}{\cellcolor{gray!15}\textbf{Real-world Repositories}} \\
Rep\_Penalty   & 0.828          & 0.461          & 0.358          & 0.395          & 62.01          & 1753               \\
Rep\_Dropout & 0 & 0.705 & 0.519 & 0.074 & 208.51 & 4096 \\
\hdashline
Greedy         & 0              & 0.738          & 0.631          & 0.297          & 203.96         & 4096          \\
Temp (t=0.1)   & 0.031          & 0.726          & 0.62           & 0.292          & 189.35         & 3997          \\
Temp (t=0.2)   & 0.052          & 0.728          & 0.627          & 0.309          & 185.38         & 3944          \\
Temp (t=0.8)   & 0.769          & 0.496          & 0.384          & 0.365          & 64.72          & 1926          \\
Topk (k=5)     & 0.732          & 0.534          & 0.427          & 0.411          & 64.58          & 1937          \\
Topk (k=10)    & 0.783          & 0.51           & 0.399          & 0.381          & 63.29          & 1911          \\
Topk (k=30)    & 0.783          & 0.495          & 0.386          & 0.372          & 64.72          & 1940          \\
Topp (p=0.8)   & 0.128          & 0.686          & 0.585          & 0.344          & 172.11         & 3728          \\
Topp (p=0.9)   & 0.221          & 0.658          & 0.55           & 0.349          & 155.82         & 3443          \\
Topp  (p=0.95) & 0.291          & 0.642          & 0.521          & 0.373          & 144.96         & 3270          \\
\hdashline
RPG (Ours)     & \textbf{0.889} & \textbf{0.416} & \textbf{0.335} & \textbf{0.638} & \textbf{60.36} & \textbf{1415} \\ \bottomrule
\end{tabular}
}
\end{table}

\section{Experimenal Results}

We systematically evaluate our approach from two main aspects. First, regarding the effectiveness of mitigating repetition problem in code generation, we conduct multi-angle evaluations on CodeRepetEval dataset: 1) We compare the performance of our RPG approach with baselines in three scenarios: Code Generation Benchmarks, Artificial Synthesis, and Real-world Repositories; 2) We valid the effect of RPG on the base LLMs across different series and sizes; 3) We explore the generalizability of RPG across different programming languages (PLs).
4) We evaluate the impact of different values of hyperparameter $\lambda$ on the performance of RPG.
5) We conduct a case study to qualitatively analyze the RPG's performance (Appendix \ref{case_study}). Second, concerning the effectiveness of code generation, we evaluate the RPG's performance compared to baselines on HumanEval(-ET) and MBPP(-ET) benchmarks.

\subsection{Repetition Mitigation}
\paragraph{Effectiveness of RPG.} As illustrated in Tables \ref{table1}, we assess our RPG approach along with various baselines on CodeRepetEval dataset. Our analysis of experimental results yields several insights: 1) \textbf{Enhancing the model's confidence in its output tends to improve the likelihood of generating repetitive sequences.} Specifically, when the hyperparameters for temperature, Topk, and Top-p sampling are set to lower values, the TR-N and TR-S metrics across the three scenarios of CodeRepetEval dataset show poorer performance, and the generation of EOS tokens becomes more challenging, thereby increasing the generation time and length of models. 
2) \textbf{The approaches for addressing content repetition in text generation are not applicable to the structural repetition problems in code generation.} 
Although Repetition Penalty can reduce TR-N and TR-S to a certain extent, they achieve this at the expense of generated code quality. Its generated code usually terminates prematurely at a position where it should not end. Repetition Dropout masks the content repetitions for self-attention during training, but it has little effect on the structural repetitions.
3) \textbf{RPG substantially outperforms other baselines on CodeRepetEval dataset, which effectively reduces repetitions and enhances the quality of the generated code.} 
RPG achieves the best performance in terms of EGP and CCP metrics on CodeRepetEval dataset across three scenarios. For the TR-N and TR-S metrics, RPG also achieves the optimal performance except for Artificial Synthesis scenario. This may be attributed to the fact that although artificially synthesized prompts tend to induce repetitions, since these prompts are not inherently natural to the model, the model more readily escapes these repetitions when sampling from a smoother distribution.

\begin{table}[ht!]
\centering
\caption{The impact of RPG using different base models on CodeRepetEval dataset across three scenarios, where AS, CGB, and RR donate Artificial Synthesis, Code Generation Benchmarks, and Real-world Repositories.}
\resizebox{0.48\textwidth}{!}{
\label{tab:my-table}
\begin{tabular}{@{}lcccccc@{}}
\toprule
\multirow{2}{*}{Approach} & \multicolumn{2}{c}{AS} & \multicolumn{2}{c}{CGB} & \multicolumn{2}{c}{RR} \\ \cmidrule(lr){2-3} \cmidrule(lr){4-5} \cmidrule(lr){6-7}
  &  \multicolumn{1}{c}{TR-S $\downarrow$}  & \multicolumn{1}{c}{CCP $\uparrow$}                & \multicolumn{1}{c}{TR-S $\downarrow$} & \multicolumn{1}{c}{CCP $\uparrow$} & \multicolumn{1}{c}{TR-S $\downarrow$} & \multicolumn{1}{c}{CCP $\uparrow$}     \\ \midrule
\multicolumn{7}{c}{\cellcolor{gray!15}\textbf{CodeLlama}} \\
 Greedy                    & 0.789               & 0.459              & 0.637                  & 0.455                 & 0.631                & 0.297                \\
 RPG              & \textbf{0.613}      & \textbf{0.731}     & \textbf{0.435}         & \textbf{0.805}        & \textbf{0.379}       & \textbf{0.638}       \\ \midrule
\multicolumn{7}{c}{\cellcolor{gray!15}\textbf{CodeGen}} \\
Greedy                    & 0.684               & 0.541              & 0.657                  & 0.509                 & 0.657                & 0.369                \\
 RPG                & \textbf{0.437}      & \textbf{0.841}     & \textbf{0.417}         & \textbf{0.867}        & \textbf{0.375}       & \textbf{0.720}       \\ \midrule
\multicolumn{7}{c}{\cellcolor{gray!15}\textbf{DeepSeek-Coder}} \\
Greedy & 0.453 & 0.821  & 0.664 & 0.479 & 0.467 & 0.332 \\
 RPG   & \textbf{0.369}  & \textbf{0.951}  & \textbf{0.515}  & \textbf{0.732} & \textbf{0.341}  & \textbf{0.585}  \\
 \midrule
\multicolumn{7}{c}{\cellcolor{gray!15}\textbf{Llama2}} \\
Greedy                    & 0.528               & 0.758              & 0.508                  & 0.597                 & 0.542                & 0.477                \\
 RPG                & \textbf{0.422}      & \textbf{0.917}     & \textbf{0.427}         & \textbf{0.912}        & \textbf{0.383}       & \textbf{0.764}       \\  \bottomrule
 
\end{tabular}}
\end{table}
\begin{figure}[th!]
\centering
\includegraphics[width=0.49\textwidth]{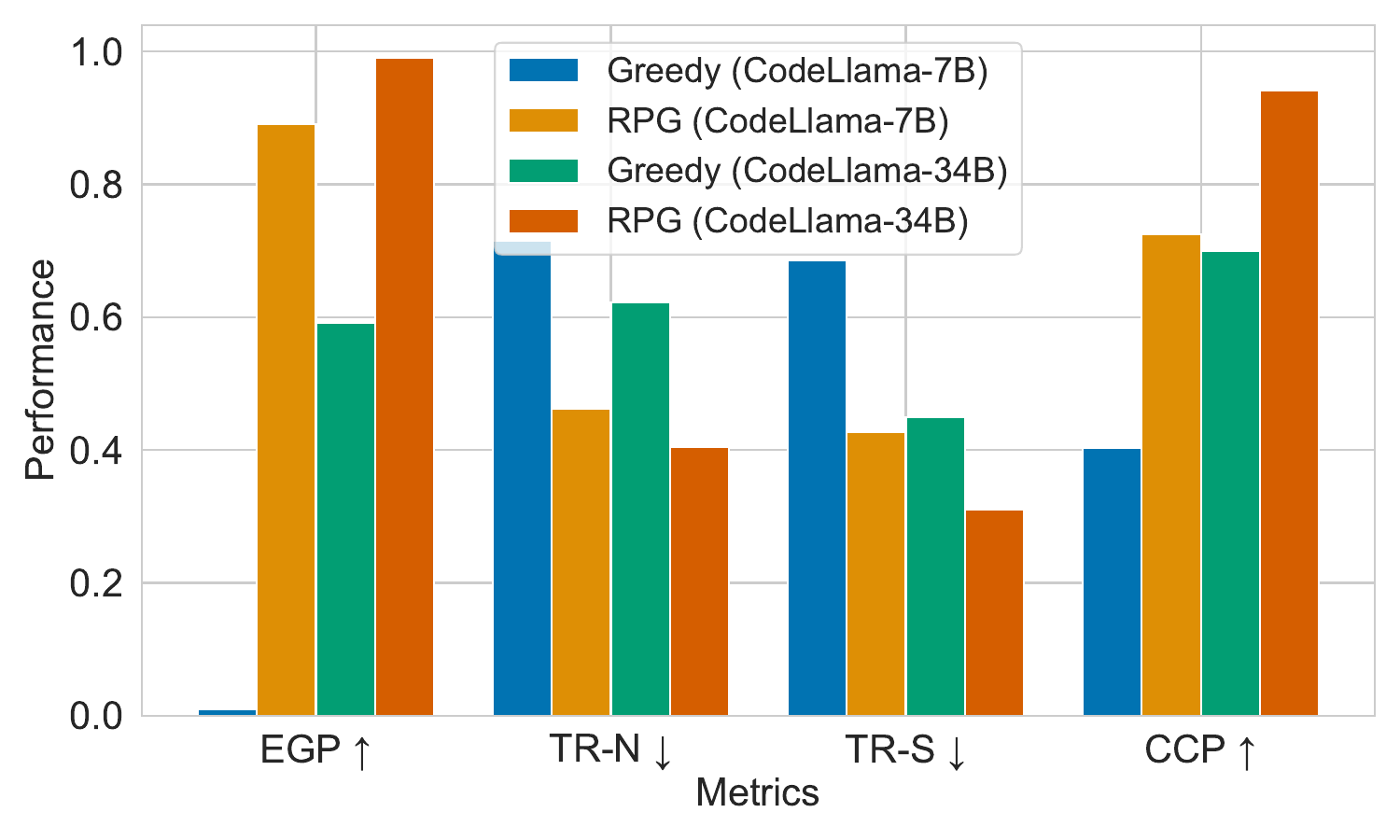}
\caption{The performance of RPG applied to LLMs of different sizes. This result is the average value across three scenarios on CodeRepetEval dataset.}
\label{diff_size}
\end{figure}

\paragraph{Performance on different LLMs.} We also conduct experiments for evaluating the performance of RPG based on different LLMs across three scenarios of CodeRepetEval dataset, as presented in Table \ref{tab:my-table}. 
The experimental results indicate that RPG exhibits consistent and significant enhancements on different base models, highlighting the robustness of RPG for base model selections. Moreover, we find that training LLMs on code is likely to increase their susceptibility to structural repetitions during greedy sampling. Given that CodeLlama, although fine-tuned on Llama2 for coding tasks, demonstrates markedly worse performance in terms of structural repetitions in code generation.

As shown in Figure \ref{diff_size}, we observe that the repetition also occurs on LLMs larger than 7B, i,e., CodeLlama-34B, and RPG is effective for it as well, which demonstrates the same trends as other LLMs evaluated in Table \ref{table1} and Table \ref{tab:my-table}, indicating the generalizability across LLMs of varying sizes.

\begin{table}[ht!]
\centering
\caption{The performance of RPG on CodeRepetEval-Go dataset.}
\resizebox{0.49\textwidth}{!}{
\begin{tabular}{lccccc}
\toprule
Approach & \multicolumn{1}{l}{EGP $\uparrow$} & \multicolumn{1}{l}{TR-N $\downarrow$} & \multicolumn{1}{l}{TR-S $\downarrow$} & \multicolumn{1}{l}{CCP $\uparrow$} & \multicolumn{1}{l}{Time $\downarrow$} \\ \midrule
Greedy & 0.133 & 0.750 & 0.353 & 0.403 & 38.28 \\
Temp (t=0.8) & 0.601 & 0.554 & 0.231 & 0.396 & 26.78 \\
RPG (Ours) & \textbf{0.875} & \textbf{0.518} & \textbf{0.215} & \textbf{0.725} & \textbf{21.07} \\
\bottomrule
\end{tabular}}\label{go_PL}
\end{table}

\paragraph{Performance on different PLs.} RPG can be applied to other PLs, requiring only their grammatical rules, which are readily obtainable from the web. To demonstrate the convenience of RPG, we have extended it to the PL of Go, and the experimental results are shown in Table \ref{go_PL}. We can find that RPG still achieved substantial improvements in all five metrics for the PL of Go.

\paragraph{Influence of hyperparamter $\lambda$.} In our experiments, we fix the hyperparameter $\lambda$ intuitively for RPG. As shown in Figure \ref{lambda} of Appendix \ref{Alambda}, we investigate the influence of varying $\lambda$ empirically on all scenarios of CodeRepetEval dataset and HumanEval and MBPP benchmarks by changing itself. The results indicate that the reduction of the hyperparameter $\lambda$ for RPG leads to a more pronounced suppression of repetition problems in code generation and decreases the sampling time. Furthermore, there is still room for further improvements with the better hyper-parameter setup of $\lambda$.

\paragraph{Case Study.} In Figure \ref{cases} of Appendix \ref{case_study}, we showcase two examples from Code Generation Benchmarks and Real-World Repository to conduct qualitative analysis. In these cases, the original model falls into a repetition trap, continuing until it exhausts its token budget. Case (a) features a repetition pattern: num = [i for i in num if i != NUM], where LLMs repeatedly generate statements that increment NUM. Case (b)'s repetition pattern involves a sequence of imports, where modules `attention' and `conv' are endlessly added. The code generated under these repetition patterns is utterly nonsensical and fails completely. However, our approach can break out of these loops, returning to a normal code generation trajectory, and ultimately succeeding in generating correct code.

\begin{table}[ht!]
\centering
\caption{The performance of RPG on code generation benchmarks HumanEval(-ET) and MBPP(-ET). Improvement represents the relative improvement of RPG compared to Greedy sampling.}
\resizebox{0.50\textwidth}{!}{
\begin{tabular}{lcccc}
\toprule
\multicolumn{1}{l}{Approach} & HumanEval & HumanEval-ET & MBPP & MBPP-ET \\ \midrule
Rep\_Penalty & 0.092 & 0.067 & 0.143 & 0.115 \\
Rep\_Dropout & 0.139 & 0.116 & 0.167 & 0.141 \\ \hdashline
Greedy & 0.301 & 0.232 & 0.396 & 0.303 \\
Temp (t=0.8) & 0.226 & 0.183 & 0.305 & 0.218 \\ \hdashline
RPG (Ours) & \textbf{0.325} & \textbf{0.258} & \textbf{0.421} & \textbf{0.334} \\
Improvement & \textbf{$\uparrow$ 8.0\%} & \textbf{$\uparrow$ 11.3\%} & \textbf{$\uparrow$ 6.4\%} & \textbf{$\uparrow$ 10.3\%} \\
\bottomrule
\end{tabular}}\label{Code_gen}
\end{table}

\subsection{Code Generation}

In addition to CodeRepetEval dataset, we further validate the effectiveness of RPG on widely used code generation benchmarks, i.e., HumanEval(-ET) and MBPP(-ET), as presented in Table \ref{Code_gen}. The results demonstrate that RPG outperforms both standard decoding approaches and specialized approaches aimed at reducing content repetition. Although Repetition Penalty and Repetition Dropout forcibly reduce content repetition in generated code, they also significantly impair the performance of code generation. In contrast, RPG not only effectively eliminates both content and structural repetition, but also enhances the accuracy of code generation for LLMs, achieving relative improvements of up to 11.3\% in Pass@1.

\section{Related Work}

\subsection{Code Generation}
Since the advent of artificial intelligence in the 1950s, code generation has been considered the Holy Grail of computer science research \citep{gulwani2017program}. With the rapid expansion of codebases and the increasing capacity of deep learning models, using deep learning for program generation has shown great potential and practicality \citep{RaychevVY14, LingBGHKWS16, WeiBolin, SunZXSMZ20, MukherjeeWCRCJ21, self-planning, Self-collaboration, DevEval, Rocode, EvoCodeBench, CodeDPO}. In recent years, the rise of pre-training techniques has brought new momentum to the field of code generation. For example, studies like CodeT5 \citep{CodeT5} and UniXcoder \citep{UniXcoder} pre-train models for code generation tasks. With the continual increase in model parameters, researchers have discovered emergent phenomena in LLMs, leading to new breakthroughs \nocite{GorM}. Against this backdrop, LLMs such as AlphaCode \citep{alphacode}, Codex \citep{codex}, CodeGeeX \citep{codegeex}, Starcoder \citep{starcoder}, CodeLlama \citep{codellama}, and DeepSeek Coder \citep{DeepSeek_Coder} have emerged.

Some work focuses on grammar-based code generation approaches \citep{TranX, Grammar_CNN, TreeBERT, CODEP}, which primarily utilize learning or decoding based on grammar rulers to enhance the grammatical correctness of generated code. However, given that all structural repetitions adhere to the grammar, i.e., they are grammatically correct, merely using grammar rules during decoding or learning grammar rules during training is not applicable to the structural repetition problem. Therefore, these approaches fail to address this problem.

\subsection{Repetition in Neural Text Generation.}
Repetition problems in neural language models have drawn increasing attention, with various interpretations and proposed solutions emerging from recent research, especially in the field of text generation \citep{Holtzman}.
Repetition Penalty \citep{Holtzman} is a commonly used approach to reduce content repetition, which prevents words or phrases that have already appeared during the generation process from being generated again. However, there are lots of key tokens that appear frequently in code generation, such as `=', `(', `[' \citep{CrystalBLEU}. Uniformly preventing these tokens in subsequent generations would be extremely detrimental to code generation. 

Previous work \citep{fu_high_flow} points out that repetition is caused by the phenomenon of self-reinforcement.
Some works address this problem during the training phase \citep{fu_high_flow, Rep22, Su}. Repetition dropout \citep{Rep23} finds a link between training data degradation and repetition, mitigating it by lowering attention to repeated words. However, compared to our RPG approach, these approaches have three primary disadvantages: 1) They require extensive training and necessitate the construction of a large amount of data for fine-tuning LLMs, which incurs substantial costs. 2) They usually hurt the code generation performance of models obviously. 3) They only focus on addressing content repetitions in text generation, without involving the prevalent issue of structural repetitions in code generation.

\section{Conclusion}
In this paper, we have formally defined structural repetition, which is the major repetition problem in code generation. We have proposed a novel decoding approach called RPG to alleviate repetition problems in code generation from grammar perspective for LLMs. By leveraging the grammar rules, RPG can recognize repetitions and strategically decay the output probability of critical tokens that contribute to repetitions. We also construct a new dataset CodeRepetEval, designed to provide a comprehensive evaluation for addressing repetition problems in code generation. Extensive experiments demonstrate the effectiveness and generalization of RPG in repetition mitigation of code generation.
Through our work, we hope to shed light on this direction and call more attention to repetition problems in code generation. 

\section{Limitations}
Our work has the following two main limitations. 

First, RPG demands slightly more computational resources than sampling to detect the repetitions. However, compared to the enormous computational overhead of LLMs, it is marginal and acceptable. 

Second, the potential reasons why LLMs induce structural repetitions in code generation remain unclear. 
Our current analysis has not touched on this aspect, which we leave for future work.

\bibliography{ref}

\newpage
\appendix
 \onecolumn

\section{Case Study}\label{case_study}

\begin{figure}[h!]
    \centering
    \begin{subfigure}[b]{\textwidth}
        \includegraphics[width=\textwidth]{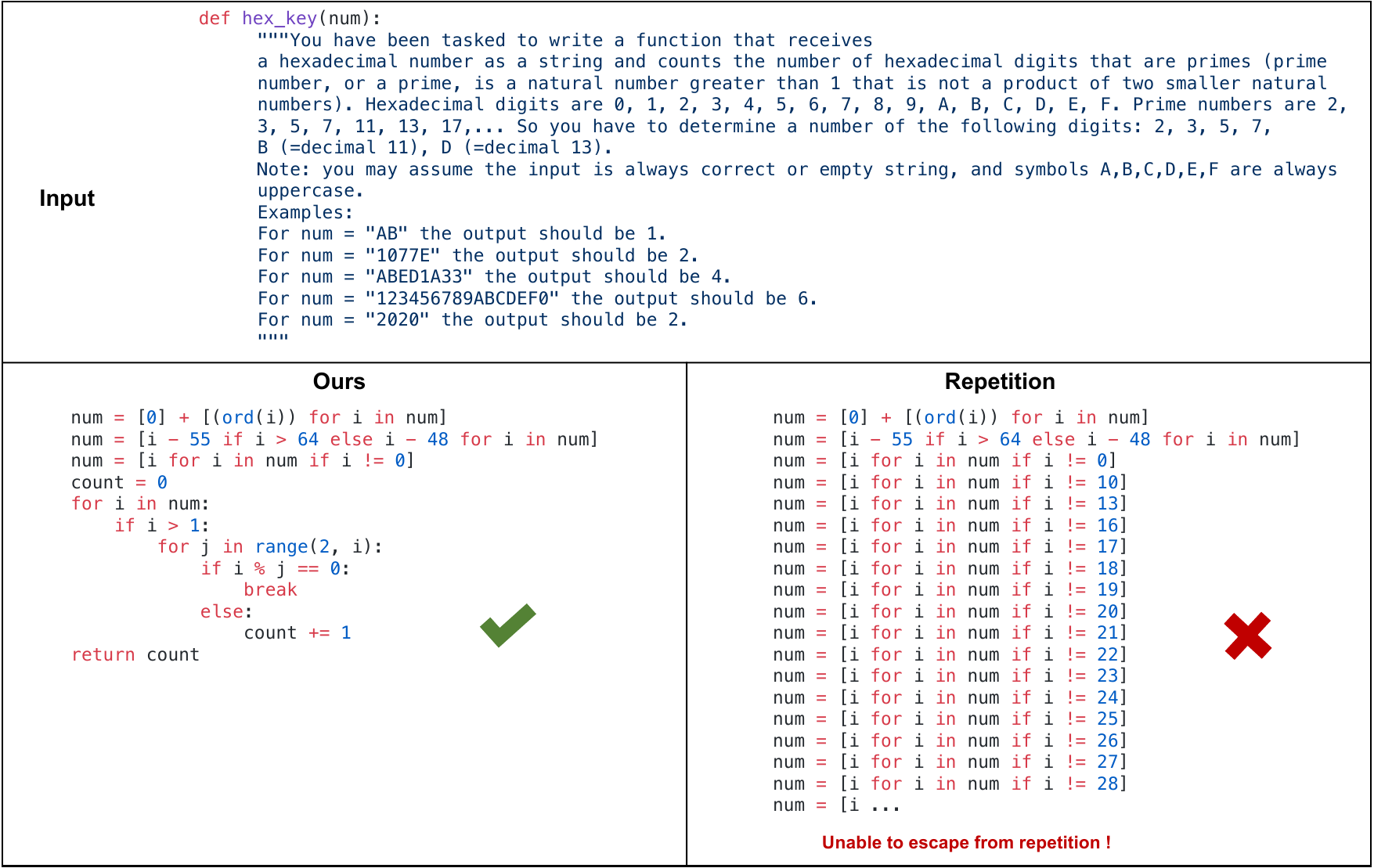}
        \caption{A case from HumanEval in Code Generation Benchmark.}
        \label{case1}
    \end{subfigure}

    \vspace{0.5cm}

    \begin{subfigure}[b]{\textwidth}
        \includegraphics[width=\textwidth]{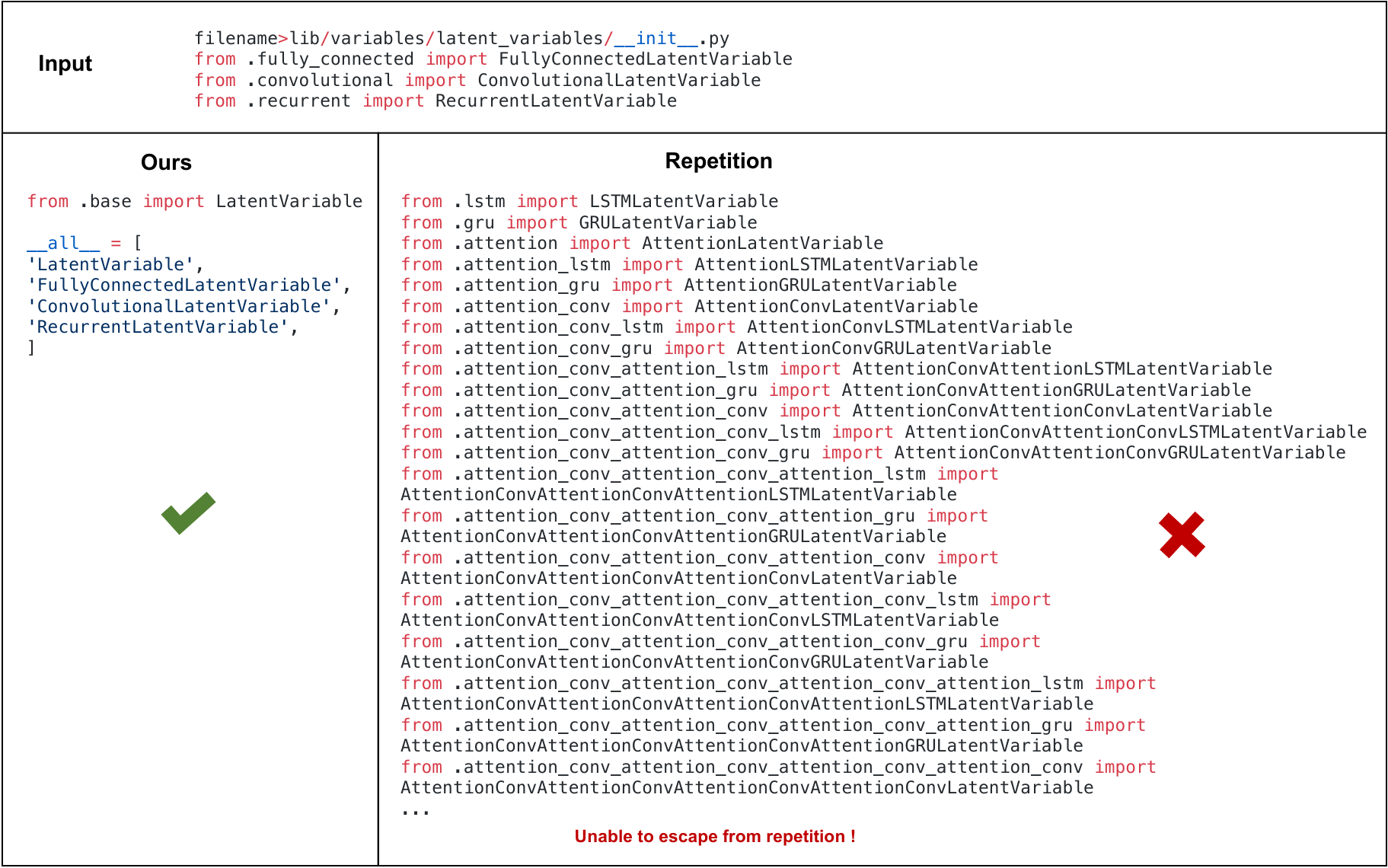}
        \caption{A case from Real-world Repository}
        \label{case2}
    \end{subfigure}
    
    \caption{Two cases of generating structural repetition and the effect of our approach on them. LLMs succumb to endless loops of repetition. Our proposed approach can effectively break out of these loops, steering back to a normal code generation trajectory, and ultimately succeeding in producing correct code.}
    \label{cases}
\end{figure}

\section{The Influence of Hyper-parameters $\lambda$}\label{Alambda}
\begin{figure*}[h!]
\centering
\includegraphics[width=1\textwidth]{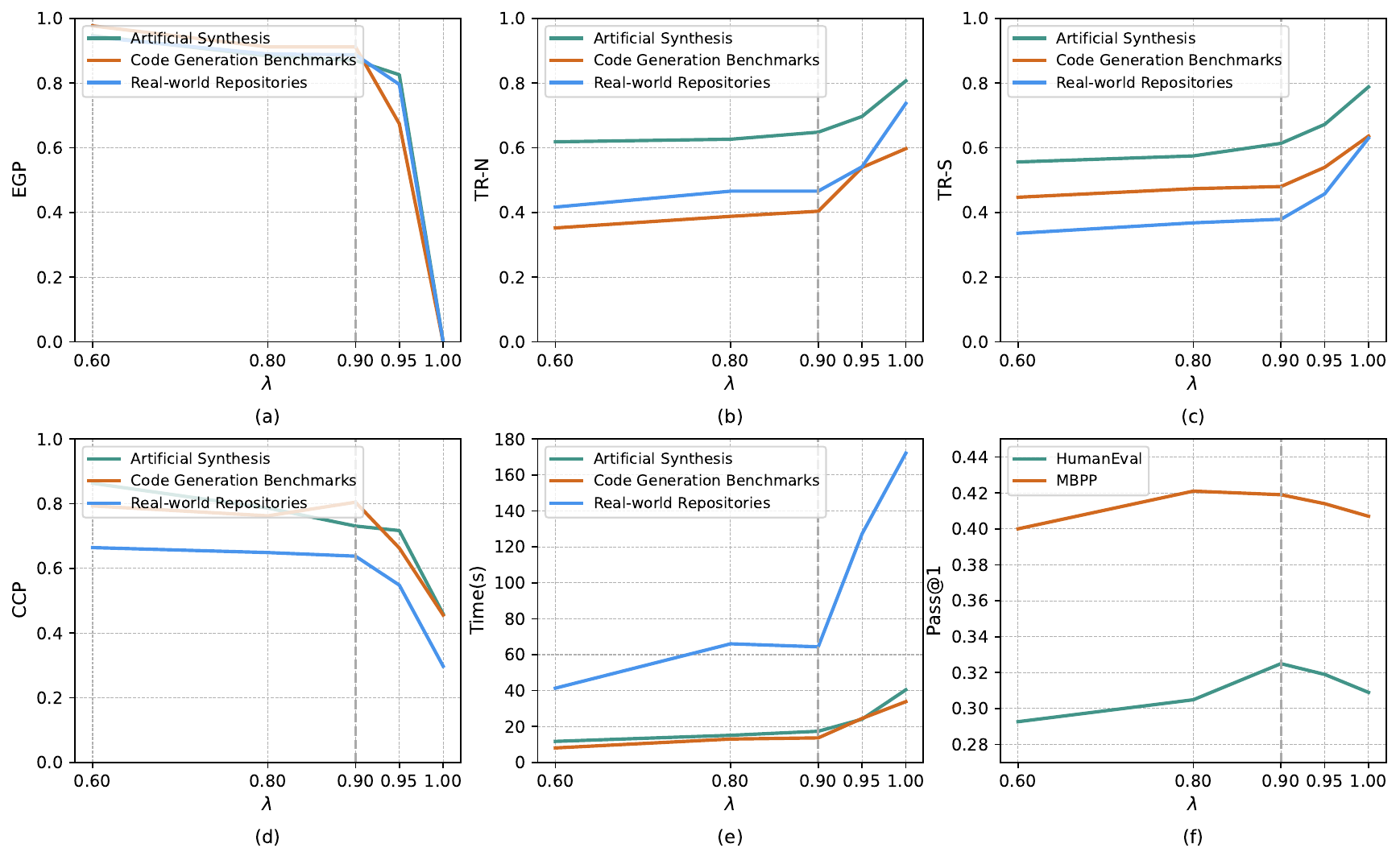}
\caption{The influence of hyper-parameters $\lambda$ on Artificial Synthesis, Code Generation Benchmarks, and Real-world Repositories scenarios of CodeRepetEval dataset, as well as HumanEval and MBPP benchmarks. We use the \textcolor{gray}{gray dashed line} to represent the employed hyper-parameters.}
\label{lambda}
\end{figure*}

We shown the influence of hyper-parameters $\lambda$ in Figure \ref{lambda}.

\section{Details of Experiment Setup}\label{app_setup}
\subsection{Dataset Construction and Evaluation Methodology of CodeRepetEval}

The purpose of CodeRepetEval datasets is to evaluate the performance of different approaches for avoiding repetitions in code generation under various scenarios. We provide a detailed description of the dataset construction process and the evaluation methodology as follows.

The construction processes for three scenarios, i.e., Artificial Synthesis, Code Generation Benchmarks, and Real-world Repositories, are similar, with mere differences in the source of code samples. For Artificial Synthesis, we employ Self-instruct~\citep{Self-instruct} to construct instruction sets that induce CodeLlama-7B to generate code samples containing repetitions. For Code Generation Benchmarks, we use CodeLlama-7B to sample codes on HumanEval and MBPP benchmarks. For Real-world Repositories, we select code from 100 high-quality open-source projects on GitHub, following the selection criteria of Starcoder~\citep{starcoder}.

After obtaining the code samples, we employ the TR-S metric to sort them in descending order based on the degree of repetition. We then select the repetitive code segments from the top 2*N samples, where N equals 512, 512, and 1024 for the respective scenarios. Subsequently, we randomly truncate the code samples at positions containing 2 to 4 repeated statements, retain the first half, and combine it with the original input to generate new prompts. Finally, the researchers conduct a manual evaluation to filter out N prompts, which are then incorporated into our datasets.

For the evaluation, we employ the constructed prompts as input and generate code using different approaches. We apply three metrics—TR-S, TR-N, and EGP—to determine the extent to which the code generated by these approaches continues to manifest repetition issues. Moreover, we apply the CCP metric to assess the compilability and quality of generated codes.

\subsection{Code Generation Benchmarks}
We evaluate our approach using four public code generation benchmarks.

\textbf{HumanEval} \citep{codex} consists of 164 handwritten programming tasks, proposed by OpenAI. Each task includes a function signature, NL description, use cases, function body, and several unit tests (average 7.7 per task).

\textbf{MBPP} \citep{mbpp} contains 974 manually verified Python programming tasks, covering programming fundamentals, standard library functionality, and more. Each task consists of an NL description, a code solution, and 3 automated test cases.

\textbf{HumanEval-ET} and \textbf{MBPP-ET} \citep{CodeScore} are expanded versions of MBPP and HumanEval respectively, where each includes over 100 additional test cases per task. This updated version enhances the soundness of code evaluation compared to the original benchmarks.

\subsection{Baselines}
\label{Baselines}

We detail the baseline approaches compared in this work:

\begin{itemize}
    \item \textbf{Greedy Sampling} chooses the highest probability token from the model's output distribution at each time step, $P_{\text{greedy}}'(w)=\mathds{1}(w=\arg\max_w P(w|w_{<t}))$. 

    \item \textbf {Topk Sampling} \citep{topk} limits the next-word selection to the top k most likely candidates as determined by the model, $P'_{\text{topk}}(w)= P(w | w_{<t}) \ \text{if } w \in \text{Topk}, \text{otherwise} \ 0$. 

    \item \textbf{Topp Sampling} \citep{Holtzman} involves choosing from a smaller set of plausible candidates by dynamically selecting a variable-sized subset of tokens (the "nucleus") that cumulatively make up a certain probability mass (e.g., top 90\%), $P'_{\text{topp}}(w) = P(w | w_{<t}) \ \text{if } \sum_{w' \in S} P(w' | w_{<t}) \leq p, \text{otherwise} \ 0$. 
    
    \item \textbf{Temperature Sampling} \citep{caccia2019language} controls the randomness of the token selection process—higher temperatures lead to more uniform distributions, while lower temperatures make high-probability tokens even more likely, $P'_{\text{temp}}(w) = \frac{\exp(\log(P(w | w_{<t})) / T)}{\sum_{w'} \exp(\log(P(w' | w_{<t})) / T)}$. 
    
    \item \textbf{Repetition Penalty}  \citep{CTRL} penalizes sampling works by discounting the scores of previously generated tokens, $P'(w) = \frac{\exp(\log(P(w | w_{<t})) / (T \cdot I(i \in g))}{\sum_{w'} \exp(\log(P(w' | w_{<t})) / (T \cdot I(i \in g))}, \ \text{where} \  I(c)=\theta \ \text{if} \ c \ \text{is True else} \ 1$. Unless otherwise specified, the settings of baselines follow their original paper.
    
    \item \textbf{Repetition Dropout} \citep{Rep23} applies masking vectors to sentences, randomly dropping out repetitive n-grams based on a pre-specified dropout rate, thereby preventing the model from over-relying on repetitive patterns during training.
\end{itemize}

\subsection{Metrics}

\paragraph{Compiler Correctness Percentage (CCP).} CCP is defined as the ratio of the number of code samples that pass compilation to the total number of samples in the dataset. 

\paragraph{Pass@k.}
We use the unbiased version \citep{codex} of Pass@k, where $n>=k$ samples are generated for each problem, count the number of correct samples $c<=n$ which pass test cases and calculate the following estimator, i.e.,
\begin{equation}
    \operatorname{Pass@k} = \mathop{\mathbb{E}}\limits_{\operatorname{Problems}}\begin{bmatrix}1-\frac{\begin{pmatrix}n-c\\k\end{pmatrix}}{\begin{pmatrix}n\\k\end{pmatrix}}\end{bmatrix}.
\end{equation}
 
\section{Pseudo-code of Suffix Array and LCP Array}\label{pseudocode}
\begin{mybox2}
\begin{lstlisting}
def constructSuffixArray(S):
    n = length of S
    Create an array suffixes[n] where each element is a tuple (index, suffix)
    for i from 0 to n-1:
        suffixes[i] = (i, S[i:n])  // Store index and suffix starting at index i
    Sort suffixes based on the suffix part of each tuple
    Initialize SA[n]
    for i from 0 to n-1:
        SA[i] = suffixes[i].index
    return SA

def constructLCPArray(S, SA):
    n = length of S
    Initialize LCP[n] with zeros
    Initialize rank[n] to store the rank of each suffix in SA
    for i from 0 to n-1:
        rank[SA[i]] = i
    h = 0  // Length of the longest common prefix
    for i from 0 to n-1:
        if rank[i] > 0:
            j = SA[rank[i] - 1]  // Index of the previous suffix in the sorted list
            while i + h < n and j + h < n and S[i + h] == S[j + h]:
                h += 1
            LCP[rank[i]] = h
            if h > 0:
                h -= 1  // Decrease h for the next calculation
    return LCP

def findConsecutiveRepetitions(S):
    SA = constructSuffixArray(S)
    LCP = constructLCPArray(S, SA)
    repetitions = set()
    for i from 1 to length of S - 1:
        if LCP[i] > 0:
            duplicate_substring = S[SA[i]:SA[i] + LCP[i]]
            # Check for consecutive occurrence
            previous_suffix_length = SA[i-1]
            current_suffix_length = SA[i]
            if previous_start == current_start + LCP[i]:
                repetitions.add(duplicate_substring)
    return repetitions
\end{lstlisting}
\end{mybox2}

\section{PDA for LLMs}
\label{PDA4LLM}
LLMs usually employ Byte-Pair Encoding (BPE) \citep{BPE} for tokenization, which causes the tokens in the vocabulary of LLMs to deviate from the terminal symbols in grammar. Specifically, this discrepancy manifests in three primary scenarios, i.e., one token corresponds to one terminal symbol, one token corresponds to multiple terminal symbols, and multiple tokens correspond to one terminal symbol. Therefore, to effectively utilize PDA with LLMs, it is essential to develop an approach that adapts PDA operations to accommodate these tokenization scenarios. 

\subsection{One Token to One Terminal Symbol}
In scenarios where one token directly corresponds to one terminal symbol, the adaptation of PDA is relatively straightforward. The PDA can process each token as a single unit that matches exactly one terminal symbol in the grammar of the language being parsed. Here, the transition functions of PDA can be directly applied without modification. 
For instance, if a token from the LLM's output matches a terminal symbol in a programming language's grammar, the PDA can push, pop, or transition based on this token following standard PDA rules. This case represents the simplest form of interaction between LLM outputs and grammar-based parsing.

\subsection{One Token to Multiple Terminal Symbols}
This scenario arises when a single token encapsulates multiple grammatical elements, due to a compact or compressed representation of the language.
For example, ``[]'', ``),'', ``)\textbackslash n'', and so on. 
To handle this, we should first decompose the token into its constituent terminal symbols. Then, we sequentially check whether each terminal symbol is in the PDA candidate set of its prefixes. Finally, the tokens that all constitute terminal symbols satisfying the condition are retained. This approach ensures that even complex tokens can be seamlessly integrated into the grammar-based processing framework of the PDA.

\subsection{Multiple Tokens to One Terminal Symbol}
In contrast, when multiple tokens collectively represent a single terminal symbol, we should aggregate these tokens before using the transition function of PDA. This scenario typically occurs with the token-types in terminal symbols, such as NAME, NUMBER, and STRING. In this case, we constructed a lexical grammar PDA to accumulate tokens until a complete terminal symbol is formed. The PDA operations then proceed based on these aggregated terminal symbols.

\section{Full Grammar specification}
\label{Full Grammar specification}
For example, the full Python grammar is shown as follows:
\begin{mybox}
\begin{lstlisting}
# Grammar for Python

# NOTE WELL: You should also follow all the steps listed at
# https://devguide.python.org/grammar/

# Start symbols for the grammar:
#       single_input is a single interactive statement;
#       file_input is a module or sequence of commands read from an input file;
#       eval_input is the input for the eval() functions.
#       func_type_input is a PEP 484 Python 2 function type comment
# NB: compound_stmt in single_input is followed by extra NEWLINE!
# NB: due to the way TYPE_COMMENT is tokenized it will always be followed by a NEWLINE
single_input: NEWLINE | simple_stmt | compound_stmt NEWLINE
file_input: (NEWLINE | stmt)* ENDMARKER
eval_input: testlist NEWLINE* ENDMARKER

decorator: '@' dotted_name [ '(' [arglist] ')' ] NEWLINE
decorators: decorator+
decorated: decorators (classdef | funcdef | async_funcdef)

async_funcdef: ASYNC funcdef
funcdef: 'def' NAME parameters ['->' test] ':' [TYPE_COMMENT] func_body_suite

parameters: '(' [typedargslist] ')'

# The following definition for typedarglist is equivalent to this set of rules:
#
#     arguments = argument (',' [TYPE_COMMENT] argument)*
#     argument = tfpdef ['=' test]
#     kwargs = '**' tfpdef [','] [TYPE_COMMENT]
#     args = '*' [tfpdef]
#     kwonly_kwargs = (',' [TYPE_COMMENT] argument)* (TYPE_COMMENT | [',' [TYPE_COMMENT] [kwargs]])
#     args_kwonly_kwargs = args kwonly_kwargs | kwargs
#     poskeyword_args_kwonly_kwargs = arguments ( TYPE_COMMENT | [',' [TYPE_COMMENT] [args_kwonly_kwargs]])
#     typedargslist_no_posonly  = poskeyword_args_kwonly_kwargs | args_kwonly_kwargs
#     typedarglist = (arguments ',' [TYPE_COMMENT] '/' [',' [[TYPE_COMMENT] typedargslist_no_posonly]])|(typedargslist_no_posonly)"
#
# It needs to be fully expanded to allow our LL(1) parser to work on it.

typedargslist: (
  (tfpdef ['=' test] (',' [TYPE_COMMENT] tfpdef ['=' test])* ',' [TYPE_COMMENT] '/' [',' [ [TYPE_COMMENT] tfpdef ['=' test] (
        ',' [TYPE_COMMENT] tfpdef ['=' test])* (TYPE_COMMENT | [',' [TYPE_COMMENT] [
        '*' [tfpdef] (',' [TYPE_COMMENT] tfpdef ['=' test])* (TYPE_COMMENT | [',' [TYPE_COMMENT] ['**' tfpdef [','] [TYPE_COMMENT]]])
      | '**' tfpdef [','] [TYPE_COMMENT]]])
  | '*' [tfpdef] (',' [TYPE_COMMENT] tfpdef ['=' test])* (TYPE_COMMENT | [',' [TYPE_COMMENT] ['**' tfpdef [','] [TYPE_COMMENT]]])
  | '**' tfpdef [','] [TYPE_COMMENT]]] )
|  (tfpdef ['=' test] (',' [TYPE_COMMENT] tfpdef ['=' test])* (TYPE_COMMENT | [',' [TYPE_COMMENT] [
   '*' [tfpdef] (',' [TYPE_COMMENT] tfpdef ['=' test])* (TYPE_COMMENT | [',' [TYPE_COMMENT] ['**' tfpdef [','] [TYPE_COMMENT]]])
  | '**' tfpdef [','] [TYPE_COMMENT]]])
  | '*' [tfpdef] (',' [TYPE_COMMENT] tfpdef ['=' test])* (TYPE_COMMENT | [',' [TYPE_COMMENT] ['**' tfpdef [','] [TYPE_COMMENT]]])
  | '**' tfpdef [','] [TYPE_COMMENT])
)
tfpdef: NAME [':' test]

# The following definition for varargslist is equivalent to this set of rules:
#
#     arguments = argument (',' argument )*
#     argument = vfpdef ['=' test]
#     kwargs = '**' vfpdef [',']
#     args = '*' [vfpdef]
#     kwonly_kwargs = (',' argument )* [',' [kwargs]]
#     args_kwonly_kwargs = args kwonly_kwargs | kwargs
#     poskeyword_args_kwonly_kwargs = arguments [',' [args_kwonly_kwargs]]
#     vararglist_no_posonly = poskeyword_args_kwonly_kwargs | args_kwonly_kwargs
#     varargslist = arguments ',' '/' [','[(vararglist_no_posonly)]] | (vararglist_no_posonly)
#
# It needs to be fully expanded to allow our LL(1) parser to work on it.

varargslist: vfpdef ['=' test ](',' vfpdef ['=' test])* ',' '/' [',' [ (vfpdef ['=' test] (',' vfpdef ['=' test])* [',' [
        '*' [vfpdef] (',' vfpdef ['=' test])* [',' ['**' vfpdef [',']]]
      | '**' vfpdef [',']]]
  | '*' [vfpdef] (',' vfpdef ['=' test])* [',' ['**' vfpdef [',']]]
  | '**' vfpdef [',']) ]] | (vfpdef ['=' test] (',' vfpdef ['=' test])* [',' [
        '*' [vfpdef] (',' vfpdef ['=' test])* [',' ['**' vfpdef [',']]]
      | '**' vfpdef [',']]]
  | '*' [vfpdef] (',' vfpdef ['=' test])* [',' ['**' vfpdef [',']]]
  | '**' vfpdef [',']
)
vfpdef: NAME

stmt: simple_stmt | compound_stmt
simple_stmt: small_stmt (';' small_stmt)* [';'] NEWLINE
small_stmt: (expr_stmt | del_stmt | pass_stmt | flow_stmt |
             import_stmt | global_stmt | nonlocal_stmt | assert_stmt)
expr_stmt: testlist_star_expr (annassign | augassign (yield_expr|testlist) |
                     [('=' (yield_expr|testlist_star_expr))+ [TYPE_COMMENT]] )
annassign: ':' test ['=' (yield_expr|testlist_star_expr)]
testlist_star_expr: (test|star_expr) (',' (test|star_expr))* [',']
augassign: ('+=' | '-=' | '*=' | '@=' | '/=' | '%=' | '&=' | '|=' | '^=' |
            '<<=' | '>>=' | '**=' | '//=')
# For normal and annotated assignments, additional restrictions enforced by the interpreter
del_stmt: 'del' exprlist
pass_stmt: 'pass'
flow_stmt: break_stmt | continue_stmt | return_stmt | raise_stmt | yield_stmt
break_stmt: 'break'
continue_stmt: 'continue'
return_stmt: 'return' [testlist_star_expr]
yield_stmt: yield_expr
raise_stmt: 'raise' [test ['from' test]]
import_stmt: import_name | import_from
import_name: 'import' dotted_as_names
# note below: the ('.' | '...') is necessary because '...' is tokenized as ELLIPSIS
import_from: ('from' (('.' | '...')* dotted_name | ('.' | '...')+)
              'import' ('*' | '(' import_as_names ')' | import_as_names))
import_as_name: NAME ['as' NAME]
dotted_as_name: dotted_name ['as' NAME]
import_as_names: import_as_name (',' import_as_name)* [',']
dotted_as_names: dotted_as_name (',' dotted_as_name)*
dotted_name: NAME ('.' NAME)*
global_stmt: 'global' NAME (',' NAME)*
nonlocal_stmt: 'nonlocal' NAME (',' NAME)*
assert_stmt: 'assert' test [',' test]

compound_stmt: if_stmt | while_stmt | for_stmt | try_stmt | with_stmt | funcdef | classdef | decorated | async_stmt
async_stmt: ASYNC (funcdef | with_stmt | for_stmt)
if_stmt: 'if' namedexpr_test ':' suite ('elif' namedexpr_test ':' suite)* ['else' ':' suite]
while_stmt: 'while' namedexpr_test ':' suite ['else' ':' suite]
for_stmt: 'for' exprlist 'in' testlist ':' [TYPE_COMMENT] suite ['else' ':' suite]
try_stmt: ('try' ':' suite
           ((except_clause ':' suite)+
            ['else' ':' suite]
            ['finally' ':' suite] |
           'finally' ':' suite))
with_stmt: 'with' with_item (',' with_item)*  ':' [TYPE_COMMENT] suite
with_item: test ['as' expr]
# NB compile.c makes sure that the default except clause is last
except_clause: 'except' [test ['as' NAME]]
suite: simple_stmt | NEWLINE INDENT stmt+ DEDENT

namedexpr_test: test [':=' test]
test: or_test ['if' or_test 'else' test] | lambdef
test_nocond: or_test | lambdef_nocond
lambdef: 'lambda' [varargslist] ':' test
lambdef_nocond: 'lambda' [varargslist] ':' test_nocond
or_test: and_test ('or' and_test)*
and_test: not_test ('and' not_test)*
not_test: 'not' not_test | comparison
comparison: expr (comp_op expr)*
# <> isn't actually a valid comparison operator in Python. It's here for the
# sake of a __future__ import described in PEP 401 (which really works :-)
comp_op: '<'|'>'|'=='|'>='|'<='|'<>'|'!='|'in'|'not' 'in'|'is'|'is' 'not'
star_expr: '*' expr
expr: xor_expr ('|' xor_expr)*
xor_expr: and_expr ('^' and_expr)*
and_expr: shift_expr ('&' shift_expr)*
shift_expr: arith_expr (('<<'|'>>') arith_expr)*
arith_expr: term (('+'|'-') term)*
term: factor (('*'|'@'|'/'|'%'|'//') factor)*
factor: ('+'|'-'|'~') factor | power
power: atom_expr ['**' factor]
atom_expr: [AWAIT] atom trailer*
atom: ('(' [yield_expr|testlist_comp] ')' |
       '[' [testlist_comp] ']' |
       '{' [dictorsetmaker] '}' |
       NAME | NUMBER | STRING+ | '...' | 'None' | 'True' | 'False')
testlist_comp: (namedexpr_test|star_expr) ( comp_for | (',' (namedexpr_test|star_expr))* [','] )
trailer: '(' [arglist] ')' | '[' subscriptlist ']' | '.' NAME
subscriptlist: subscript (',' subscript)* [',']
subscript: test | [test] ':' [test] [sliceop]
sliceop: ':' [test]
exprlist: (expr|star_expr) (',' (expr|star_expr))* [',']
testlist: test (',' test)* [',']
dictorsetmaker: ( ((test ':' test | '**' expr)
                   (comp_for | (',' (test ':' test | '**' expr))* [','])) |
                  ((test | star_expr)
                   (comp_for | (',' (test | star_expr))* [','])) )

classdef: 'class' NAME ['(' [arglist] ')'] ':' suite

arglist: argument (',' argument)*  [',']

# The reason that keywords are test nodes instead of NAME is that using NAME
# results in an ambiguity. ast.c makes sure it's a NAME.
# "test '=' test" is really "keyword '=' test", but we have no such token.
# These need to be in a single rule to avoid grammar that is ambiguous
# to our LL(1) parser. Even though 'test' includes '*expr' in star_expr,
# we explicitly match '*' here, too, to give it proper precedence.
# Illegal combinations and orderings are blocked in ast.c:
# multiple (test comp_for) arguments are blocked; keyword unpackings
# that precede iterable unpackings are blocked; etc.
argument: ( test [comp_for] |
            test ':=' test |
            test '=' test |
            '**' test |
            '*' test )

comp_iter: comp_for | comp_if
sync_comp_for: 'for' exprlist 'in' or_test [comp_iter]
comp_for: [ASYNC] sync_comp_for
comp_if: 'if' test_nocond [comp_iter]

# not used in grammar, but may appear in "node" passed from Parser to Compiler
encoding_decl: NAME

yield_expr: 'yield' [yield_arg]
yield_arg: 'from' test | testlist_star_expr

# the TYPE_COMMENT in suites is only parsed for funcdefs,
# but can't go elsewhere due to ambiguity
func_body_suite: simple_stmt | NEWLINE [TYPE_COMMENT NEWLINE] INDENT stmt+ DEDENT

func_type_input: func_type NEWLINE* ENDMARKER
func_type: '(' [typelist] ')' '->' test
# typelist is a modified typedargslist (see above)
typelist: (test (',' test)* [','
       ['*' [test] (',' test)* [',' '**' test] | '**' test]]
     |  '*' [test] (',' test)* [',' '**' test] | '**' test)
\end{lstlisting}
\end{mybox}

\end{document}